\documentclass[runningheads]{llncs}
\usepackage{graphicx}
\usepackage{caption}
\usepackage{subcaption}
\usepackage{float}
\usepackage[table,xcdraw]{xcolor}
\begin{document}
\title{AKHCRNet: Bengali handwritten character recognition using deep learning}
%
%

\author{Roy, Akash}
\authorrunning{Roy, Akash}
%
\institute{Government College of Engineering and Ceramic Technology
\email{royakashappleid@icloud.com}}
\maketitle              
\begin{abstract}
I propose a state of the art deep neural architectural solution for handwritten character recognition for Bengali alphabets, compound characters as well as numerical digits that achieves state-of-the-art accuracy 96.8\% in just 11 epochs. Similar work has been done before by Chatterjee, Swagato, et al.\cite{kingshuk chatterhejee} but they achieved 96.12\% accuracy in about 47 epochs. The deep neural architecture used in that paper was fairly large considering the inclusion of the weights of the ResNet 50 model which is a 50 layer Residual Network. This proposed model achieves higher accuracy as compared to any previous work \& in a little number of epochs. ResNet50 is a good model trained on the ImageNet dataset, but I propose an HCR network that is trained from the scratch on Bengali characters without the "Ensemble Learning" that can outperform previous architectures.

\end{abstract}
\section{Introduction}
\subsection{Motivation}
Bengali is one of the official languages of the Republic of India and the official language of Bangladesh. About 230 million people speak and write Bengali as a native language around the world. So recognition of Bengali character is an important problem needed to be solved which could potentially help solving other noble applications like optical character recognition, handwritten character recognition and even word recognition.\\

\noindent But solving Bengali character recognition is much more tough than the English counterpart because 
\begin{itemize}
    \item There are far more characters in Bengali alphabets,
    \item Bengali has conjunct consonants in it's alphabets which are harder to classify because it consists of two distinct Bengali characters. Any learning algorithm could simply classify those characters as one or the other,
    \item Many Bengali characters resembles each other closely maybe differentiated by small lines and dots, for example 1st and the 2nd vowels in Bengali alphabets are differentiated by one vertical line.
\end{itemize}

\noindent The problem of English character recognition may have been solved but the Bengali counterpart still have not achieved the human level accuracy.

\noindent Although there is a substantial amount of work that has been done on this topic as detailed in the previously done section improvements still can be achieved as the benchmark has not yet reached human-level accuracy or surpassed that. Using recent advancements both in the usage of convolution kernels in Deep Learning and also in model training procedures (like hyper-parameter tuning, data augmentation, etc.) can increase the accuracy of HCR Models. Although scientists who most recently worked on this topic used the transfer learning from the ResNet50 model achieved 96.12\% accuracy, I used a sophisticated architecture and trained from scratch \& got a better result. The BanglaLekha-Isolated Dataset as described in datasets section was used for evaluation mainly because it has a large sample size and large variance and also because the output classes are balanced. The dataset has a total of 84 characters, which has 10 numerals, 50 basic characters, and 24 most common conjunct-consonants.

\subsection{Previous works}
Several other people laid the foundation for Bengali handwritten character recognition. In those, work Ray and Chatterjee did the first significant work\cite{firstwork} using nearest neighbour classifiers. After that, many more researchers did other methods, improving the performance of Bangla Handwritten Character Recognition (HCR) like Hasant et al.\cite{hasant} proposed an HCR capable of classifying both printed and handwritten characters by using Discrete Cosine Transform method over the input image and Hidden Markov Model (HMM) for character classification. Wen et al.\cite{wen} proposed a Bangla numerals recognition method using Principal Component Analysis and Support Vector Machines, but those are using traditional machine learning methods. In Hassan and Khan’s work, they used the KNN algorithm where features were extracted using local binary patterns. Das et al.\cite{das} proposed a feature set representation for Bangla handwritten character recognition. It was done by hand engineering 8 distance features, 24 shadow features, 84 quad-tree based longest run features, and 16 centroid features. Their accuracy was about 85.40\% on a 50 character class dataset as hand engineering was difficult for the Conjunct-Consonants in Bengali characters. Rumman Rashid Chowdhury used CNN based classifier\cite{paper-Rumman} and gained over 98\% accuracy on 10 neumericals and 91.12\% accuracy on 50 basic Bengali character. In the mentioned methods, the main concern is that they used many handcrafted features extracted for the small dataset, and also those didn't achieved the human level accuracy. So these methods are in no means achieved the human level accuracy in handwritten character recognition. As convolutional neural networks eliminated the need for feature extraction, it'll be much more economical to make the HCR network to learn the features on it's own.

\section{Dataset}

The dataset I’m going to be using for this problem is the BanglaLekha Isolated Dataset\cite{data}. There are many open datasets that could be used for this classification problem but this dataset is chosen because of the followings:
\begin{itemize}
    \item This dataset has over 166,000 examples of Bengali characters. Neural networks can learn better with huge amounts of data.
    \item Dataset is made with male as well as female native Bengali speakers of different states of Bangladesh. So the dataset is not biased towards a specific gender category.
    \item Their age ranging from 4-27. So the dataset is not biased to a certain age groups
    \item This dataset also introduces handwritten data from the people of special ability. So the dataset is not biased to a people of special abilities.
    \item This dataset also doesn't have a class imbalance problem (More or less same amount of examples for each category).    
\end{itemize}

But the data has some blank images or images that are not taken properly or mislabeled data due to human error. So I have manually gone through every directory of 84 classes and removed some of the mislabeled data the dataset. This process is much important as I am using an algorithm to generate the validation dataset (data the neural network hasn’t seen yet) taken at random from the directory. If any of those mislabeled or blank images get into the validation set the neural network will never know and this will generate avoidable errors in the predictions. But in the training dataset, if 1-2 images get blank or mislabeled error, it doesn’t matter much. But due to sheer amount of data and unavailability of man power I only managed to remove mislabeled data from 20 categories.

Dataset preparation goes like this: Out of 166K (after cleaning mislabeled and blank images) I’ve used 28\% of data randomly chosen from the 84 classes, so no human bias for selection of validation data is present and the rest of the images are used for training.

This is how the dataset is structured:
\begin{table}[]
\centering
\begin{tabular}{|l|r|r|}
\hline
\rowcolor[HTML]{333333} 
\cellcolor[HTML]{000000}{\color[HTML]{FFFFFF} Dataset} & \multicolumn{1}{l|}{\cellcolor[HTML]{333333}{\color[HTML]{FFFFFF} Total Images Used}} & \multicolumn{1}{l|}{\cellcolor[HTML]{333333}{\color[HTML]{FFFFFF} Percentage of total data}} \\ \hline
Training                                               & 118394                                                                                & 72\%                                                                                         \\ \hline
Validation                                             & 45983                                                                                 & 28\%                                                                                         \\ \hline
\end{tabular}
\end{table}

So the validation data is selected by computer at random before the training and the neural network doesn’t see the validation data during training, this way we can safely measure the accuracy of the network. The accuracy measuring algorithm used here is described below in the architecture section. In the next section it's shown how data is processed before going into the network.

\section{Image data pre-processing}

The very recent work that has been done by Chatterjee, Swagato, et al.\cite{kingshuk chatterhejee} achieved the best result yet for a training of 47 epochs at 96.12\% and the 97\% accuracy achieved by Saha and Saha et. al used ensemble CNN architecture run for 500 epochs. Individually, their accuracy was 95.67\% and 92.43\%.

Let’s look deeper at what Chatterjee, Swagato, et al. did. They used ResNet50 model, which is pre-trained on the ImageNet dataset. For a hand-written classification task I think that they've used transfer learning from a model that is specifically not trained to classify handwritten characters. I think that ResNet 50 model’s CNN learnt to detect several features that is not quite useful for this task of classifying hand written characters. If I’d to use a model, I'd use models that maybe trained previously on similar handwritten datasets like the MNIST dataset\cite{mnist}, I think that would converge much faster and could get better results.

Secondly they used images of size sometimes height of 200 px, width of 200 px, color depth of 3 to comply with ResNet architecture, other times 128 by 128, but in general these are huge image size. I got rid of the unnecessary color channels. I used gray-scale image as an input shape because fundamentally you wouldn’t care whether your bengali handwritten character is colorful or not in order to classify them. This made less information but the 'same' useful information going into the network, so the training will converge sooner than that of K Chatterjee et.al’s training duration which is 47 epochs.

Next step of pre-processing is to decrease the amount of information coming in neural network. I propose instead of using 200 pixels for width and height that is 40,000 parameters we should be using 32 pixels for width and height that is 1024 parameters. Let’s have a look at the data and see if those image resizing processing actually hurts the data and make them unrecognizable for the network? Each pixel value of the images is also normalized between 0-1 to converge faster, and this also makes the neural network easier to train.

I'm using the idea of bi-linear interpolation technique when resizing the input images from the original size, that goes like this:

When we are given an image suppose that we want to find the value of the unknown function f at a specific point in image (x, y). It is assumed that we know the value of function f at the four points being 
$$Q_{11} = (x_1, y_1),$$
$$Q_{12} = (x_1, y_2),$$ 
$$Q_{21} = (x_2, y_1),$$
$$Q_{22} = (x_2, y_2).$$ We first do linear interpolation in the x-direction. This gives us the following:
\begin{equation}
f(x, y_1) \approx \frac{x_2 - x}{x_2 - x_1} f(Q_{11}) + \frac{x - x_1}{x_2 - x_1} f(Q_{21})
\end{equation}

\begin{equation}
f(x, y_2) \approx \frac{x_2 - x}{x_2 - x_1} f(Q_{12}) + \frac{x - x_1}{x_2 - x_1} f(Q_{22})
\end{equation}
Then to the y direction:

\begin{equation}
f(x, y) \approx \frac{y_2- y}{y_2 - y_1} f(x, y_{1}) + \frac{y - y_1}{y_2 - y_1} f(x, y_2)
\end{equation}
$$=\frac{y_2- y}{y_2 - y_1} \left ( \frac{x_2 - x}{x_2 - x_1} f(Q_{11}) + \frac{x - x_1}{x_2 - x_1} f(Q_{21}) \right ) + \frac{y - y_1}{y_2 - y_1} \left ( \frac{x_2 - x}{x_2 - x_1} f(Q_{12}) + \frac{x - x_1}{x_2 - x_1} f(Q_{22}) \right )$$

\begin{figure}[H]
    \centering
    \includegraphics[width=\textwidth]{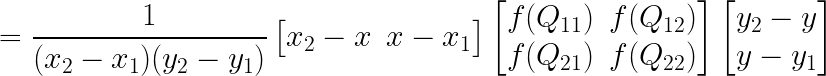}
    \label{fig:matrix}
\end{figure}
Using this bi-linear interpolation technique for image resizing you can see the result here:

\begin{figure}[H]
\centering
\begin{subfigure}{0.5\textwidth}
  \centering
  \includegraphics[width=.4\linewidth]{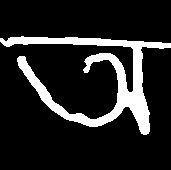}
  \caption{Image Size (171, 170)}
  \label{fig:1}
  \end{subfigure}%
\begin{subfigure}{0.5\textwidth}
  \centering
  \includegraphics[width=.4\linewidth]{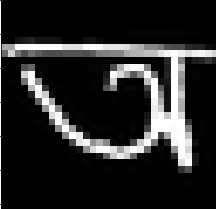}
  \caption{Image Size (32, 32)}
  \label{fig:2}
\end{subfigure}
\caption{Class 1 Image before resize and after}
\label{fig:test}
\end{figure}

So we can clearly see that our images are still recognizable and so we can get away with 1024 pixels, which will not going to hurt the deep learning model for less data in an image. Let's see more images resized to (32, 32, 1) and how those looks:

\begin{figure}[H]
\centering
\begin{subfigure}{0.5\textwidth}
  \centering
  \includegraphics[width=.4\linewidth]{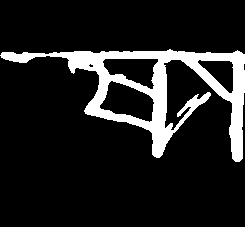}
  \caption{Image Size (245, 227)}
  \label{fig:3}
  \end{subfigure}%
\begin{subfigure}{0.5\textwidth}
  \centering
  \includegraphics[width=.4\linewidth]{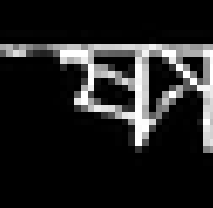}
  \caption{Image Size (32, 32)}
  \label{fig:4}
\end{subfigure}
\caption{Class 80 image resized before \& after}
\label{fig:test}
\end{figure}

\section{Architecture}
The data is coming into the network as (32, 32) gray-scale image, so I have added a (32, 32, 1) layer as the input layer. For the next two layers I've used the (5, 5) convolutions and 32 filters of those, this combination worked great against a single (3, 3) convolution. These convolution layers have "same" padding as we have very little information on the images and we don't want to lose those during convolutions. Throughout the entire deep neural architecture, the ReLU activation is used. Where $$ReLU(z) = max(0, z)$$ Next, a Batch-Normalization is used, this makes our model's weights of deeper layers more robust to changes than to the first few layers.  Next, I've added a Max Pooling Layer.

I've added a block of (1, 1) convolution with 128 filters followed by one (3, 3), one (5, 5) convolution with 128 filters separately and one (1, 1) convolution and one "same" padded max-pooling layer followed by a (3, 3) convolution. All convolution layers mentioned here have "same" padding. Without the inclusion of this block the architecture converged to a 96.38\% accuracy in 20 epochs, which still has the better validation accuracy than that of previous state of the art architecture, but with the inclusion of this block validation accuracy peaked 96.80\% in just 11 epochs.\\

\noindent The total of 4 outputs from the previous block has been concatenated into a (16, 16, 448) layer and a ReLU activation is applied on top of it. Following this there is another two sets of convolution-maxpooling-batch norm-maxpooling blocks. In these two blocks the convolutions has 256 and 512 filters respectively. Same padding and ReLU activation is used throughout and the data is flattened out in the next layer.\\

\noindent Following this I've added fully connected layer which is 4 layer deep and each layer has 1024, 512, 256, 128 number of hidden units respectively and the output layer has 84 units for the 84 classes.\\

\begin{figure}
    \centering
    \includegraphics[width=\textwidth]{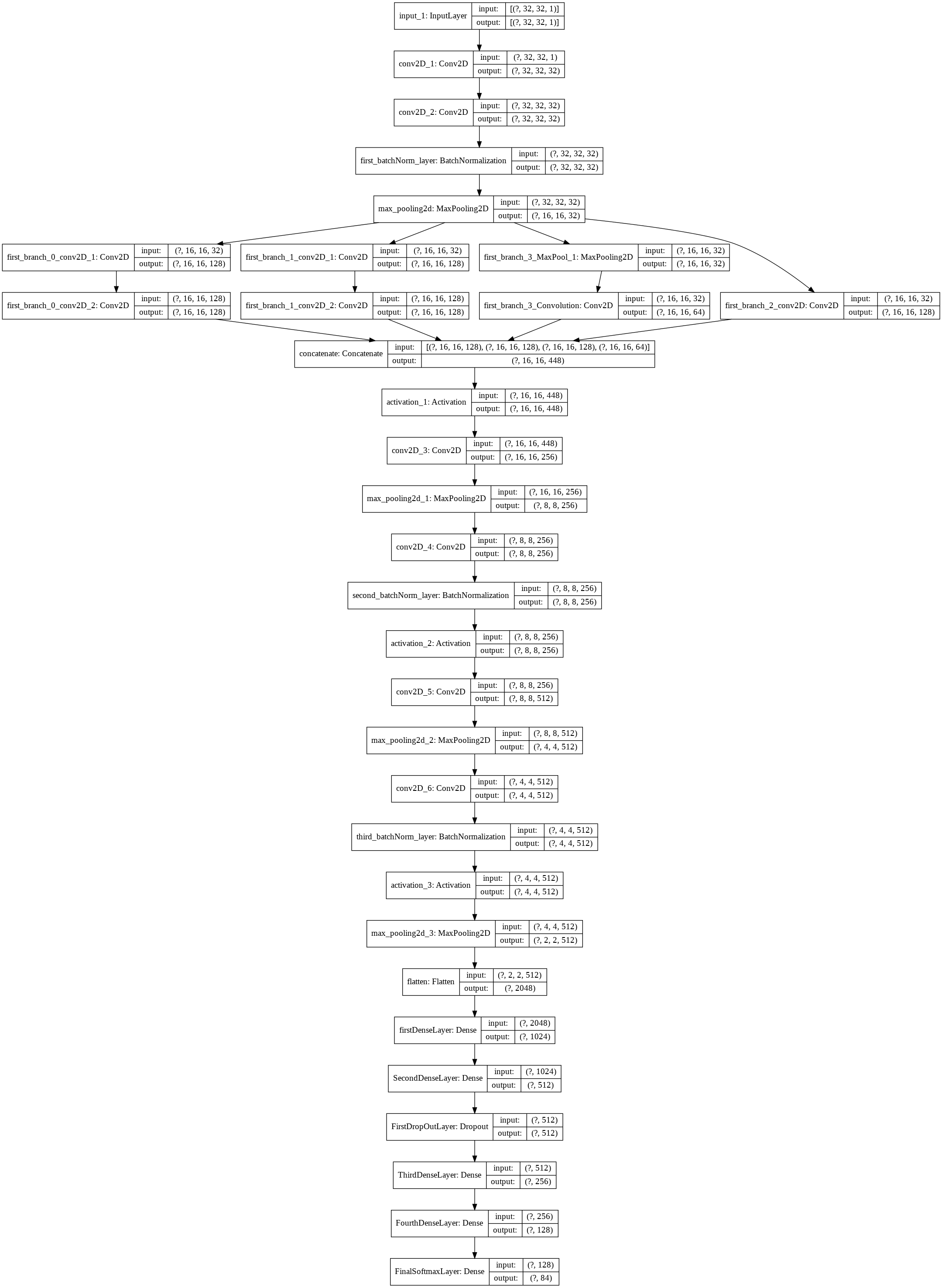}
    \caption{Layer by layer image data flow in AKHCRNet Model}
    \label{fig:my_label}
\end{figure}

\noindent The last layer has 84 hidden units with Softmax activation applied to it to comply with the 84 predictable classes in the Bengali dataset. The last few hidden layers are penalized with L2 Regularization method to decrease overfitting, Cortes, Corinna et. al~\cite{l2paper}. The regularizers used in the last few hidden layers are kernel regularizer, as weights in the network is a very high dimensional parameter-vector, but the bias is just a single real number. so almost all the parameters are in weights rather than in the bias. So I think adding a bias regularizer won't make that much of a difference. So it's useful to only use the kernel regularizer in this problem. 

For a L layer deep network the loss function J is the following: $$J(w^{[1]}, b^{[2]}, ... , w^{[L]}, b^{[L]})$$ Where W[i], b[i] is the i-th layers weight vector and bias respectively. The function J is a softmax cross-entropy loss also known as categorical cross-entropy loss shown below. Softmax function:

\begin{equation}
    f(s)_i = \frac{e^{s_i}}{\sum_{j}^{C} e^{s_{j}}}
\end{equation}
Categorical cross-entropy loss:
\begin{equation}
    CCE= -\sum_{i}^{C} t_i \log(f(s)_i)
\end{equation}

In this specific case of Multi-Class classification the labels are one-hot, so only the positive class Cp  keeps its position in the loss. There is only one element of the target vector t which is not zero ti = tp. So discarding the elements of the summation which are zero due to target labels, we can say:
$$CCE= -\log\left ( \frac{e^{s_p} }{\sum_{j}^{C} e^{s_j}}\right)$$
Here ti and si are the ground truth and the CNN prediction score for each class i in all classes C, An softmax activation is applied.

To make understanding clear from now on the categorical loss will be represented as $$ L(\hat{y}, y)$$ where y hat is the predicted label from softmax output unit for a given ground truth y. After adding the L2 regularization the loss function J has become:

$$ J(w^{[1]}, b^{[2]}, ... , w^{[L]}, b^{[L]}) = \frac{1}{m} \sum_{1}^{n} L(\hat{y}, y) + \frac{\lambda}{2m} \sum_{1}^{L} {\left \| W^{[L]} \right \|}^2$$
The extra term $$\frac{\lambda}{2m} \sum_{1}^{L} {\left \| W^{[L]} \right \|}^2$$ is the L2 weight regularizer. W is a weight vector of dimensions (n[L-1], n[L]) where n[L-1] and n[L] is the number of hidden unit in the L-1 th and Lth layer of the deep network. So the weight decay is now applied so that overfitting doesn't occur. The reason behind this deep network is to ensure not to under-fit the network. I used a 7-10 layer deep network, it was getting 90\% to 92\% accurate in 20 epochs. So adding more layers has made the architecture more prone to learning intricate features that it hasn't learnt before.

As this is turning out to be a very deep network I also added a dropout with the rate of 50\% in the second hidden layer. Experiments are made and it is seen that between 20-80\% dropout rate of 50\% worked the best for this problem.

\section{Training, Learning rate scheduling, Optimization}

The Adam Optimizer algorithm is used for this architecture, with no change to the default hyper-parameters $$\alpha = 1e-3$$
$$\beta_1 = 0.9$$
$$\beta_2 = 0.999$$
$$\epsilon = 1e-8$$ and instead of using a automated learning rate scheduler I've manually tweaked the learning rate epoch by epoch. Here is how that goes:

\begin{table}[H]
\centering
\begin{tabular}{|
>{\columncolor[HTML]{656565}}l |r|r|r|}
\hline
\cellcolor[HTML]{000000}{\color[HTML]{FFFFFF} Epochs} & \cellcolor[HTML]{333333}{\color[HTML]{FFFFFF} Set Learning rate} & \cellcolor[HTML]{333333}{\color[HTML]{FFFFFF} Accuracy achieved} & \cellcolor[HTML]{343434}{\color[HTML]{FFFFFF} Cross Entropy Loss} \\ \hline
{\color[HTML]{FFFFFF} First 5 Epochs}                 & 0.001                                                            & 95.11\%                                                          & 0.30081                                                           \\ \hline
{\color[HTML]{FFFFFF} Next 3 Epochs}                  & 0.0001                                                           & 96.79\%                                                          & 0.22147                                                           \\ \hline
{\color[HTML]{FFFFFF} Last 3 Epochs}                  & 0.00004                                                          & 96.80\%                                                          & 0.21612                                                           \\ \hline
\end{tabular}
\label{tab:my-table}
\end{table}

The result after 11 epochs is shown in the Result Section.

\section{Result}
In the following figure Figure-4 epoch by epoch validation loss and accuracy is shown. Clearly no sign of overfitting can be seen, the overall accuracy is 96.80\% in 11 epochs which is a monumental jump from previous 500 epoch CNN ensemble training by Saha and Saha et al. and 47 epoch training by K Chatterjee et al.

\begin{figure}[H]
\centering
\begin{subfigure}{0.5\textwidth}
  \centering
  \includegraphics[width=\linewidth]{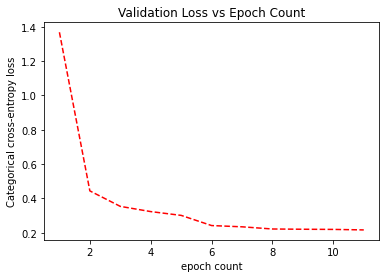}
  \caption{Validation loss over time}
  \label{fig:5}
  \end{subfigure}%
\begin{subfigure}{0.5\textwidth}
  \centering
  \includegraphics[width=\linewidth]{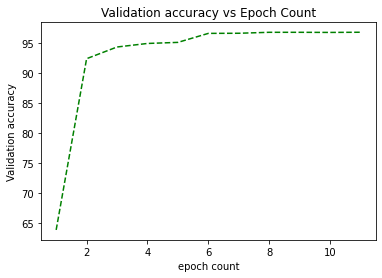}
  \caption{Validation accuracy over time}
  \label{fig:6}
\end{subfigure}
\caption{validation accuracy and loss with each passing iteration.}
\label{fig:test}
\end{figure}

Below is the confusion matrix of 84 classes of Bengali character predictions. You can also see all the precision, recall, f1-score for each category in the following link. \texttt{https://bit.ly/akhcrnetreport-csv}

\begin{figure}[H]
\centering
\includegraphics[width=5cm]{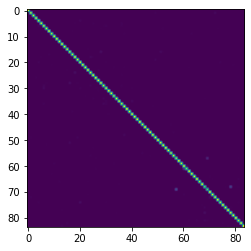}
\caption{Confusion Matrix}
\label{fig:6}
\end{figure}

Here Precision is the ratio of correctly predicted positive observations in Bengali characters to the total positive predictions. The question that this metric answer is of all characters that labeled as some label, how many of them is actually that label? The high precision always relates to the low false positive rate. We have got about 0.97 (avg over 84 class examples) precision which is pretty good for a hand character recognizer. The report for each class can be seen following the link \texttt{https://bit.ly/akhcrnetreport-csv}.

$$precision = \frac{TP}{(TP+FP)}$$

Recall (Sensitivity) - Recall is the ratio of correctly predicted positive observations to the all observations in actual class - yes. The question recall answers is: Of all the characters that truly that label, how many did we label? We have got recall of 0.97 which is good for this model.

$$recall = \frac{TP}{(TP+FN)}$$

F1 score - F1 Score is calculated by the weighted average of precision and recall. Therefore, this score takes into account both false positives and false negatives. Intuitively it is not as easy to understand as accuracy, but f1-score can be more useful than accuracy, as we have an some-what uneven class distribution. Accuracy works best when the false positives and false negatives have the similar cost. If the cost of false positives and false negatives are very different, it’s better to look at both Precision and Recall. In our case, F1 score is 0.97 (avg). The f1-score for each class can be seen following the link 

\texttt{https://bit.ly/akhcrnetreport-csv}.

$$f1-score = \frac{2*(Recall * Precision)} { (Recall + Precision)}$$

\subsection{Discussion}
I call this architecture AKHCRNet which is a fancy way of saying Hand Character recognizer Network by by Akash. This architecture out-performs every other proposed Bengali classifier algorithms. Let's see how other researchers works did back then:
\begin{table}[H]
\centering
\begin{tabular}{|
>{\columncolor[HTML]{343434}}l |r|l|r|r|}
\hline
{\color[HTML]{FFFFFF} \begin{tabular}[c]{@{}l@{}}Researchers\\ and their work\end{tabular}} & \cellcolor[HTML]{333333}{\color[HTML]{FFFFFF} \begin{tabular}[c]{@{}r@{}}Total \\ Classes\end{tabular}} & \cellcolor[HTML]{333333}{\color[HTML]{FFFFFF} Architecture}                                    & \cellcolor[HTML]{343434}{\color[HTML]{FFFFFF} Accuracy}                          & \cellcolor[HTML]{333333}{\color[HTML]{FFFFFF} Epochs} \\ \hline
{\color[HTML]{FFFFFF} Purkayastha et al.\cite{Purkayastha et al.}}                                                   & 50                                                                                                      & Vanilla CNN                                                                                    & 89.01\%                                                                          &                                                       \\ \hline

{\color[HTML]{FFFFFF} Rahmen et al.\cite{Rahman et al.}}                                                        & 50                                                                                                      & CNN                                                                                            & 85.96\%                                                                          &                                                       \\ \hline
{\color[HTML]{FFFFFF} Alif et al.\cite{Alif et al}}                                                          & 84                                                                                                      & \begin{tabular}[c]{@{}l@{}}Modified ResNet-18\\ with dropout layer of\\ 20\% rate\end{tabular} & 95.10\%                                                                          &                                                       \\ \hline
{\color[HTML]{FFFFFF} \begin{tabular}[c]{@{}l@{}}Rumman Rashid\\ Chowdhury et al.\cite{paper-Rumman}\end{tabular}}    & \begin{tabular}[c]{@{}r@{}}50 Characters\\+ 10 Numbers\end{tabular}                                        & Custom CNN                                                                                     & \begin{tabular}[c]{@{}r@{}}91.13 \% (50 Char)\\ + 98.42\% (Numbers)\end{tabular} & 92                                                     \\ \hline
{\color[HTML]{FFFFFF} Majid and Smith\cite{Majid et al.}}                                                      & 50                                                                                                      & \begin{tabular}[c]{@{}l@{}}Non CNN\\ Architecture\end{tabular}                                 & 96.80\%                                                                          &                                                       \\ \hline
{\color[HTML]{FFFFFF} Saha and Saha\cite{saha}}                                                        & 84                                                                                                      & Ensemble CNN                                                                                   & \begin{tabular}[c]{@{}r@{}}97.21\%\\ (95.67\% \& \\ 96.12\%)\end{tabular}        & 500                                                   \\ \hline
{\color[HTML]{FFFFFF} \begin{tabular}[c]{@{}l@{}}Chatterjee,\\ Swagato, et al.\cite{kingshuk chatterhejee}\end{tabular}}   & 84                                                                                                      & \begin{tabular}[c]{@{}l@{}}Transfer Learning on\\ ResNet-50\end{tabular}                       & 96.12\%                                                                          & 47                                                    \\ \hline
{\color[HTML]{FFFFFF} Proposed Work}                                                        & 84                                                                                                      & \begin{tabular}[c]{@{}l@{}}Custom 32-Layer Deep\\ Network, AKHCRNet\end{tabular}               & 96.80\%                                                                          & 11                                                    \\ \hline
\end{tabular}
\label{tab:researchers table}
\end{table}

The Saha and Saha paper did the ensemble of CNN, with two CNN architecture, each of which had the accuracy 95.67\% and 92.43\% respectively, but AKHCRNet model was able to achieve the accuracy of 96.80\%. The code and the weight file is available for download here:

\texttt{https://www.github.com/theroyakash/AKHCRNet}. 

\section{The Conclusion and Future Scope}

AKHCRNet model was able to achieve the accuracy of 96.80\% on the BanglaLekha-Isolated Dataset of 166K images, Without ensembling, without transfer learning and with a minimal number of epochs our proposed solution achieves state of the art result on Bengali Handwritten character recognition.\\

\noindent AKHCRNet's custom very deep neural architecture trains on a hard and large dataset from scratch without transfer learning and achieves a 96.8\% accuracy with just 11 epochs. The success of this architecture could make possible other indic {\sl (Indic is an adjective that refers to the Indo-Aryan languages)} character classification easier.

\end{document}